\documentclass{svmult}

\usepackage{makeidx}     
\usepackage{graphicx}    
\usepackage{multicol}    
\usepackage{multirow} 
\usepackage{url}

\makeindex             

\begin{document}

\authorrunning{A. da Silva, Y. Lechevallier, F. Rossi and F. de Carvalho} 

\title*{Clustering Dynamic Web Usage Data}
\author{Alzennyr da Silva\inst{1} \and Yves Lechevallier\inst{1} \and Fabrice Rossi\inst{1} \and
Francisco de Carvalho\inst{2}}
\institute{
Project AxIS, INRIA-Rocquencourt\\ Domaine de Voluceau, B.P. 105 \\ 
78153 Le Chesnay cedex, France \\ \texttt{\{Alzennyr.Da\_Silva,Yves.Lechevallier,Fabrice.Rossi\}@inria.fr } \\
\and 
Centro de Informatica - CIn / UFPE\\
Av. Prof. Luiz Freire, s/n, CDU \\ 50740-540 Recife, Brazil\\ 
\texttt{fatc@cin.ufpe.br}
}
%
%
%
\maketitle

Most classification methods are based on the assumption that data conforms to a stationary distribution. The machine learning domain currently suffers from a lack of classification techniques that are able to detect the occurrence of a change in the underlying data distribution. Ignoring possible changes in the underlying concept, also known as concept drift, may degrade the performance of the classification model. Often these changes make the model inconsistent and regular updatings become necessary. Taking the temporal dimension into account during the analysis of Web usage data is a necessity, since the way a site is visited may indeed evolve due to modifications in the structure and content of the site, or even due to changes in the behavior of certain user groups. One solution to this problem, proposed in this article, is to update models using summaries obtained by means of an evolutionary approach based on an intelligent clustering approach. We carry out various clustering strategies that are applied on time sub-periods. To validate our approach we apply two external evaluation criteria which compare different partitions from the same data set. Our experiments show that the proposed approach is efficient to detect the occurrence of changes.

\section{Introduction}
\label{sec:1}

Web Mining \cite{kosala00web} appeared at the end of 90s and consists in
using Data Mining techniques in order to develop methods
that allow relevant information to be extracted from Web
data (such as documents, interaction traces, link structure,
etc.). A more specialized branch of this domain, called Web
Usage Mining (WUM) \cite{Cooley1999} \cite{spiliopoulou99data}, deals with techniques based on Data
Mining that are applied to the analysis of users behavior in
a website. The present article is placed is this last
context. 

WUM consists in extracting interesting information from files which register Web usage traces. This activity has become very critical for effective website management. In the e-commerce domain, for example, one of the most important
motivations for the analysis of usage is the need to build up consumer loyalty and to make the site more appealing to
new visitors. Accurate Web usage information could help to attract new customers, retain current customers, improve cross marketing/sales, measure the effectiveness of promotional campaigns, track leaving customers and find the most effective logical structure for their Web space. Other applications we can cite are the creation of adaptive websites, support services, personalization, network traffic flow analysis, etc. 

Most traditional methods in this domain take into account the entire period during which usage traces were recorded, the results obtained naturally being those which prevail over the total period. Consequently, certain types of behaviors, which take place during short sub-periods are not detected and thus remain undiscovered by traditional methods. It is, however, important to study these behaviors and thus carry out an analysis related to significant time sub-periods. It will then be possible to study the temporal evolution of users' profiles by providing descriptions that can integrate the temporal aspect. The access patterns to Web pages are indeed of a dynamic nature, due both to the on-going changes in the content and structure of the website and to changes in the users' interest. The access patterns can be influenced by certain parameters of a temporal nature, such as the time of the day, the day of the week, recurrent factors (summer/winter vacations, national holidays, Christmas) and non-recurrent global events (epidemics, wars, economic crises, the World Cup). Furthermore, as the volume of mined data is great, it is important to define summaries to represent user profiles.

WUM has just recently started to take account of temporal dependence in usage patterns. In \cite{RoddickSpiliopoulou2002}, the authors survey the work to date and explore the issues involved and the outstanding problems in temporal data mining by means of a discussion about temporal rules and their semantic. In addition, they investigate the confluence of data mining and temporal semantics. Recently in \cite{Laxman2006}, the authors outline methods for discovering sequential patterns, frequent episodes and partial periodic patterns in temporal data mining. They also discuss techniques for the statistical analysis of such approaches. Notwithstanding these considerations, the majority of methods in WUM are applied over the entire period that covers all the available data. Consequently, these methods reveal the most predominant behaviors in data, and the interesting short-term behaviors which may occur during short periods of time are not taken into account. For example, when the data analysed is inserted into a dynamic domain related to a potential long period of time (such as in the case of Web log files), it is to be expected that behaviors evolve over time. 

These considerations have given rise to many studies in data analysis, especially concerning the adaptation of traditional static data-based methods to the dynamic data framework. In this line of research, our proposition is to use summaries obtained by an evolutionary clustering approach applied over time sub-periods to carry out a follow-up of the user profile evolution. The main objective of this article is to propose and evaluate a monitoring clustering strategy which is able to find changes in a stream of Web usage data.

This article is organized as follows. Section 2 describes the proposed clustering approach based on time sub-periods. Section 3 presents the experimental framework including the benchmark data set analysed, the algorithm and the external evaluation criteria adopted as well as a discussion on the results obtained. The last section reports the final conclusion and some suggestions for future work.

\section{Clustering approach based on time sub-periods}
\label{sec:clusteringapproach}

In unsupervised classification modelling of dynamic data, new clusters may emerge at any time and existing clusters may evolve or disappear. The clustering problem can be seen as an evolutionary process of detecting and tracking dynamic clusters.  
The approach proposed in this article consists initially in splitting the entire time period analysed into more significant sub-periods. In our experiments, we use the months of the year to define the time sub-periods (cf. section \ref{sec:resulys}). This is done with the aim of discovering the evolution of old patterns or the emergence of new ones. This fact would not have been revealed by a global analysis over the whole time period. After this first step, a clustering method is applied on the data of each time sub-period, as well as over the complete period. The results provided for each clustering are then compared. 

The main base of our proposition regards the specification of a compact structure to summarize the clustering of data within a time sub-period without storing the entire data set. This structure is designated by the cluster centre (prototype) and represents a user profile. These prototypes reflect the behaviour of individuals belonging to the same cluster.

The following subsections describe in detail the four clustering strategies we study. 

\subsection{Global clustering}
\begin{figure*}
\centering
\includegraphics[width=1\textwidth]{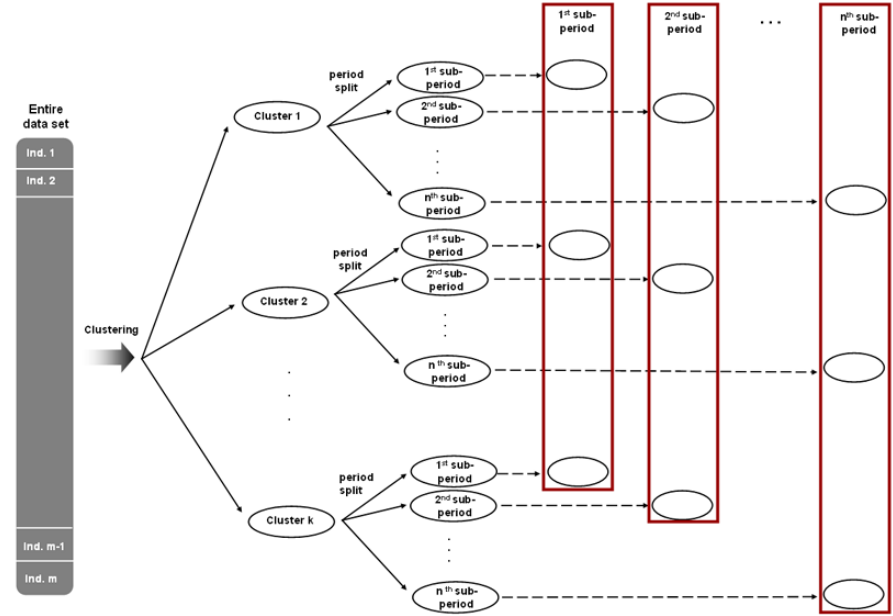}
\caption{Global clustering} \label{fig_global_clustering}
\end{figure*}

This clustering strategy corresponds to the traditional practice, that is to say, the clustering algorithm is applied on the entire data set containing all the individuals, without taking into account the temporal information. After applying the clustering algorithm, we have a partition containing K clusters. We then apply a filter on each of these clusters in order to define new sub-clusters containing individuals belonging to the same time sub-periods (for example, the same month of the year) (cf. figure \ref{fig_global_clustering}). After that we put together sub-clusters belonging to the same time sub-periods, which defines one partition per sub-period. We will then compare these last partitions with those obtained by the other clustering strategies.

\subsection{Independent local clustering}
\begin{figure*}
\centering
\includegraphics[width=0.6\textwidth,height=0.35\textheight]{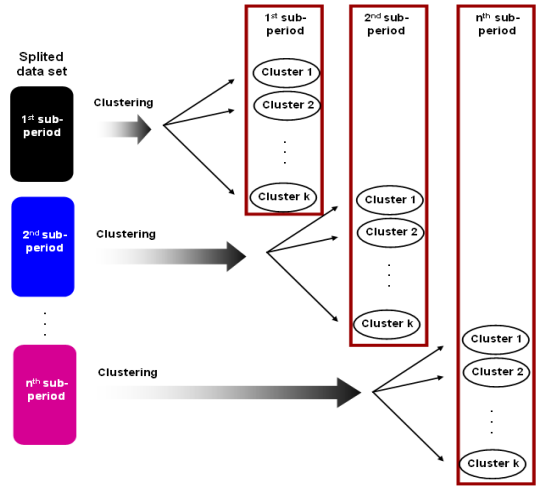}
\caption{Independent local clustering} \label{fig_independent_local_clustering}
\end{figure*}

In this clustering strategy, we first split the data set by time sub-period. After that, we apply the clustering algorithm on each sub-period individually, which gives us as many partitions as there are time sub-periods. At the end of this process we will have a partition containing K clusters in each time sub-period, each independent from the other. (cf. figure \ref{fig_independent_local_clustering}).

\subsection{Previous local clustering}
\label{previous_local_clustering}

\begin{figure*}
\centering
\includegraphics[width=1\textwidth,height=0.35\textheight]{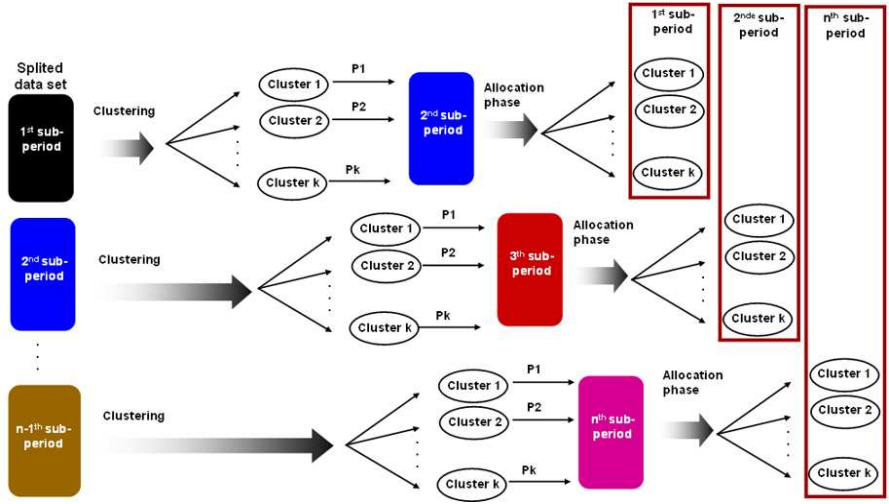}
\caption{Previous local clustering} \label{fig_previous_local_clustering}
\end{figure*}

In this clustering strategy, we begin by applying the clustering algorithm in the first time sub-period, which gives us a partition into K clusters. We take the prototype of each of these clusters in order to define a new partition on the individuals belonging to the next time sub-period. To do so, we apply the clustering allocation phase. We repeat this process for all the following time sub-periods (cf. figure \ref{fig_previous_local_clustering}).

\subsection{Dependent local clustering}
\label{dependent_local_clustering}

This clustering strategy has some points in common with the previous one. The main difference here is that we use the prototypes obtained by the clustering on the previous time sub-periods to initiate a completely new clustering process rather than applying only the allocation phase. In other words, we run the algorithm until its convergence.

\begin{figure*}
\centering
\includegraphics[width=1\textwidth,height=0.32\textheight]{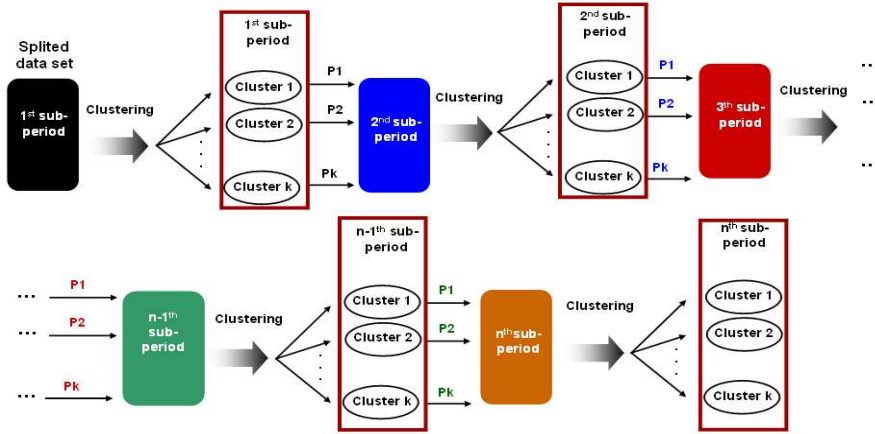}
\caption{Dependent local clustering} \label{fig_dependent_local_clustering}
\end{figure*}	

\section{Experimental framework}
\label{sec:experiments}

\subsection{Usage data}

As a case study, we use a benchmark website from Brazil \footnote{This website is available at the following address: \texttt{http://www.cin.ufpe.br/}}. This site contains a set of static pages (details of teaching staff, academic courses, etc.) and dynamic pages (see \cite{DaSilva2006}\cite{GrC06}\cite{RossiEtAlEGC2006}\cite{RossiEtAlIFCS2006} for an analysis of this part of the site). We studied the accesses to the site from 1$^{st}$ July 2002 to 31$^{st}$ May 2003.

For Web usage data pre-processing, we adopt the methodology proposed by \cite{Tanasa:2004:IEEE-IS} who defines a \textit{navigation} as a succession of requests not more than 30 minutes apart, coming from the same user. 
In order to analyse the more representative traces of usage, we selected long navigations (containing at least 10 requests and with a total duration of at least 60 seconds) which are assumed to have originated from human users (the ratio between the duration and number of requests must be at least 4, which means a maximum of 15 requests per minute). This was done in order to extract human navigations and exclude those which may well have come from Web robots. The elimination of short navigations is justified by the search for usage patterns in the site rather than simple accesses which do not generate a trajectory in the site. 
After filtering and eliminating outliers, we obtained a total of 138,536 navigations.

\begin{table}
\centering
\caption{Description of the variables describing navigations} \label{tab_champ}
\vspace{2pt}
\label{tb:variables}
\small
\begin{tabular}{c@{\quad}ll}
\hline
\multicolumn{1}{c}{\rule{0pt}{12pt}N$^{o}$} & \multicolumn{1}{l}{Field} & \multicolumn{1}{l}{Signification} \\[2pt]
\hline
   1 & IDNavigation        	& Navigation code        \\ 
   2 & NbRequests\_OK       & Number of successful requests (status = 200) in the navigation        \\ 
   3 & NbRequests\_BAD      & Number of failed requests (status $\neq$ 200) in the navigation          \\ 
   4 & PRequests\_OK        & Percentage of successful requests ( = NbRequests\_OK/ NbRequests)        \\ 
   5 & NbRepetitions        & Number of repeated requests in the navigation        \\ 
   6 & PRepetitions					& Percentage of repetitions ( = NbRepetitions / NbRequests)                \\ 
   7 & TotalDuration 				&	Total duration of the navigation (in seconds)               \\ 
   8 & AvDuration						& Average of duration ( = TotalDuration / NbRequests)              \\
   9 & AvDuration\_OK				& Average of duration among successful requests \\ 
   	 & 											& ( = TotalDuration\_OK/NbRequests\_OK)               \\ 
   10 & NbRequests\_SEM 		& Number of requests related to pages in the site's semantic structure     \\ 
   11 & PRequests\_SEM			& Percentage of requests related to pages in the site's semantic structure \\ 
      & 										& (=NbRequests\_Sem/ NbRequests)              \\ 
   12 & TotalSize						& Total size of transferred bytes in the navigation              \\ 
   13 & AvTotalSize					& Average of transferred bytes ( = TotalSize / NbRequests\_OK)             \\ 
   14 & MaxDuration\_OK			& Duration of the longest request in the navigation    (in seconds)        \\ [2pt]
\hline
\end{tabular}
\end{table}

\subsection{Clustering algorithm}

Our method uses an adapted version of the dynamic clustering algorithm \cite{Anderberg1973}\cite{DS76} \cite{mac67} applied on a data table containing the navigations in its rows and real-value variables in its columns (cf. table \ref{tb:variables}). As a distance measure, we adopt the Euclidean distance. For all the experiments, we defined an \textit{a priori} number of clusters equal to 10 with a maximum number of iterations equal to 100. The number of random initialisations is equal to 100, except when the algorithm is initialised with the results obtained from a previous execution (for the strategies presented in section \ref{previous_local_clustering} and section \ref{dependent_local_clustering}).

\subsection{Evaluation criteria}

To analyse the results, we apply two external criteria. This implies that we evaluate the results of a clustering algorithm based on how well the clustering matches pre-specified gold standard classes. For a cluster-by-cluster analysis, we compute the F-measure \cite{VanRijsbergen1979Fmesure}. To compare two partitions, we look for the best representation of the cluster $a$ in the first partition by a cluster $b$ in the second partition, i.e., we look for the best match between the clusters of two given partitions. This gives us as many values as there are clusters in the first partition. 

The F-measure combines in a single value the measures of precision $P$ and recall $R$ between an \textit{a priori} partition containing $C$ clusters and the partition reached by the clustering method containing $Q$ clusters .

\begin{figure}[h]
	\centering
		\includegraphics[width=0.5\textwidth]{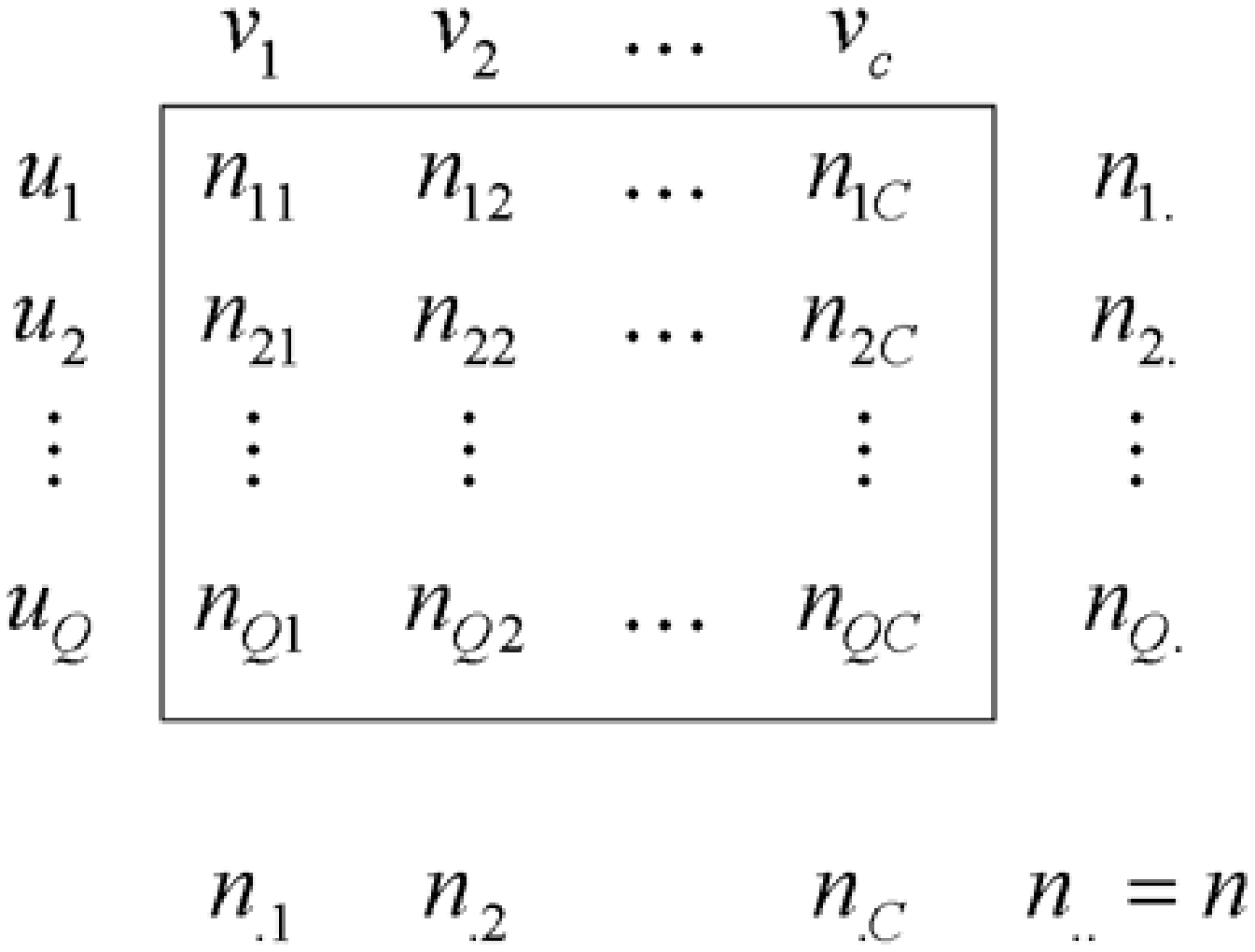}
	\caption{Contingency table}
	\label{fig:contingencytable}
\end{figure}

Let $V=\{v_1,\dots,v_c,\dots,v_C\}$ be the \textit{a priori} partition and $U=\{u_1,\dots,u_q,$ $\dots,u_Q\}$ the partition reached by the clustering method. Their contingency table is depicted in figure \ref{fig:contingencytable}. The F-measure is then defined as follows:

\begin{eqnarray}
F &=& \sum^{C}_{i=1}\frac{n_{i}}{n}max_{j=1,\ldots,Q}F(i,j) \mbox{, where } \\
F(i,j) &=& \frac{2P(i,j)R(i,j)}{P(i,j)+R(i,j)} \nonumber 
\end{eqnarray}

In this formula, $n_{i}$ indicates the number of elements in the cluster $v_i$ and $n$ is the total number of elements.

For a global analysis, we apply the corrected Rand (CR) index \cite{huba85} to compare two partitions. The CR index also assesses the degree of agreement between an \textit{a priori} partition and a partition furnished by the clustering algorithm. We use the CR index because it is not sensitive to the number of clusters in the partitions or to the distributions of the individuals in the clusters. The CR is defined as:

\begin{equation} 
\label{eq:adjustedrand} \mbox{CR} = \frac{\sum\limits_{i=1}^Q\sum\limits_{j=1}^C {{n_{ij}}\choose{2}} -
{{n}\choose{2}}^{-1}\sum\limits_{i=1}^Q{{n_{i.}}\choose{2}}\sum\limits_{j=1}^C{{n_{.j}}\choose{2}}}
{\frac{1}{2}[\sum\limits_{i=1}^Q{{n_{i.}}\choose{2}}+\sum\limits_{j=1}^C{{n_{.j}}\choose{2}}]
-{{n}\choose{2}}^{-1}\sum\limits_{i=1}^Q{{n_{i.}}\choose{2}}\sum\limits_{j=1}^C{{n_{.j}}\choose{2}}}
\end{equation}

where ${n\choose{2}}=\frac{n(n-1)}{2}$, $n_{ij}$ represents the number of objects that are in clusters $u_{i}$ and $v_{j}$, $n_{i.}$ indicates the number of objects in cluster $u_{i}$, $n_{.j}$ indicates the number of objects in cluster $v_{j}$ and $n$ is the total number of objects in the data set.

The F-measure takes a value in the range [0,+1], whereas the corrected Rand index values are in the range [-1,+1]. In both cases, the value 1 indicates a perfect agreement and values near 0 correspond to cluster agreements found by chance. In fact, an analysis made by \cite{Milligan_Cooper_86} confirmed corrected Rand index values near 0 when presented to clusters
generated from random data, and showed that values lower than 0.05 indicate clusters achieved by chance.

\subsection{Results}
\label{sec:resulys}

\begin{figure*}
\centering
\includegraphics[width=1\textwidth,height=4in]{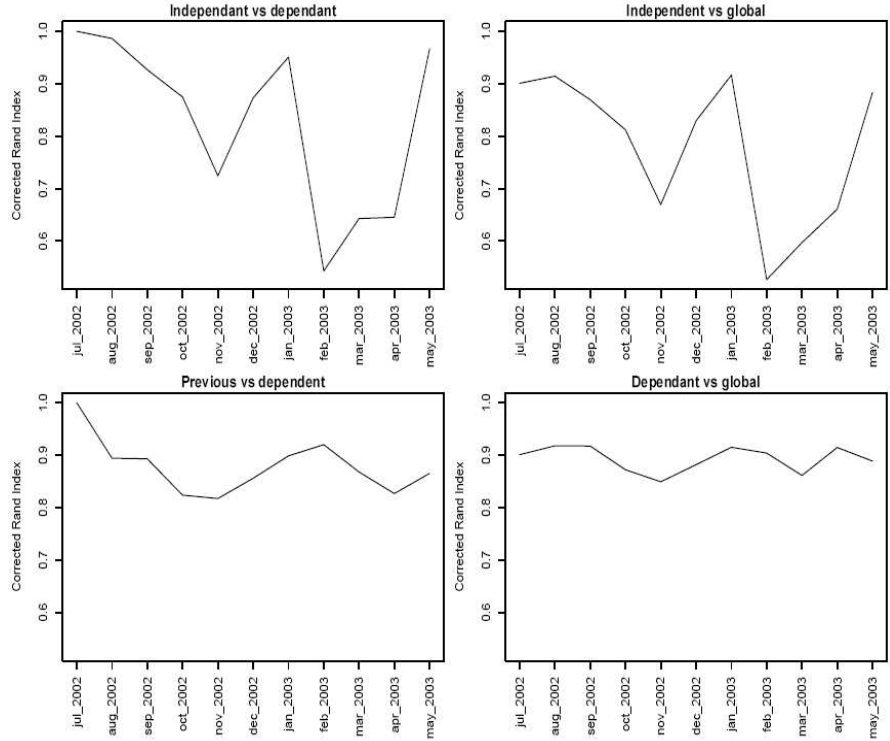}
\caption{Corrected Rand index values computed partition by partition} \label{fig_rand}
\end{figure*}

\begin{figure*}
\centering
\includegraphics[width=0.99\textwidth,height=4in]{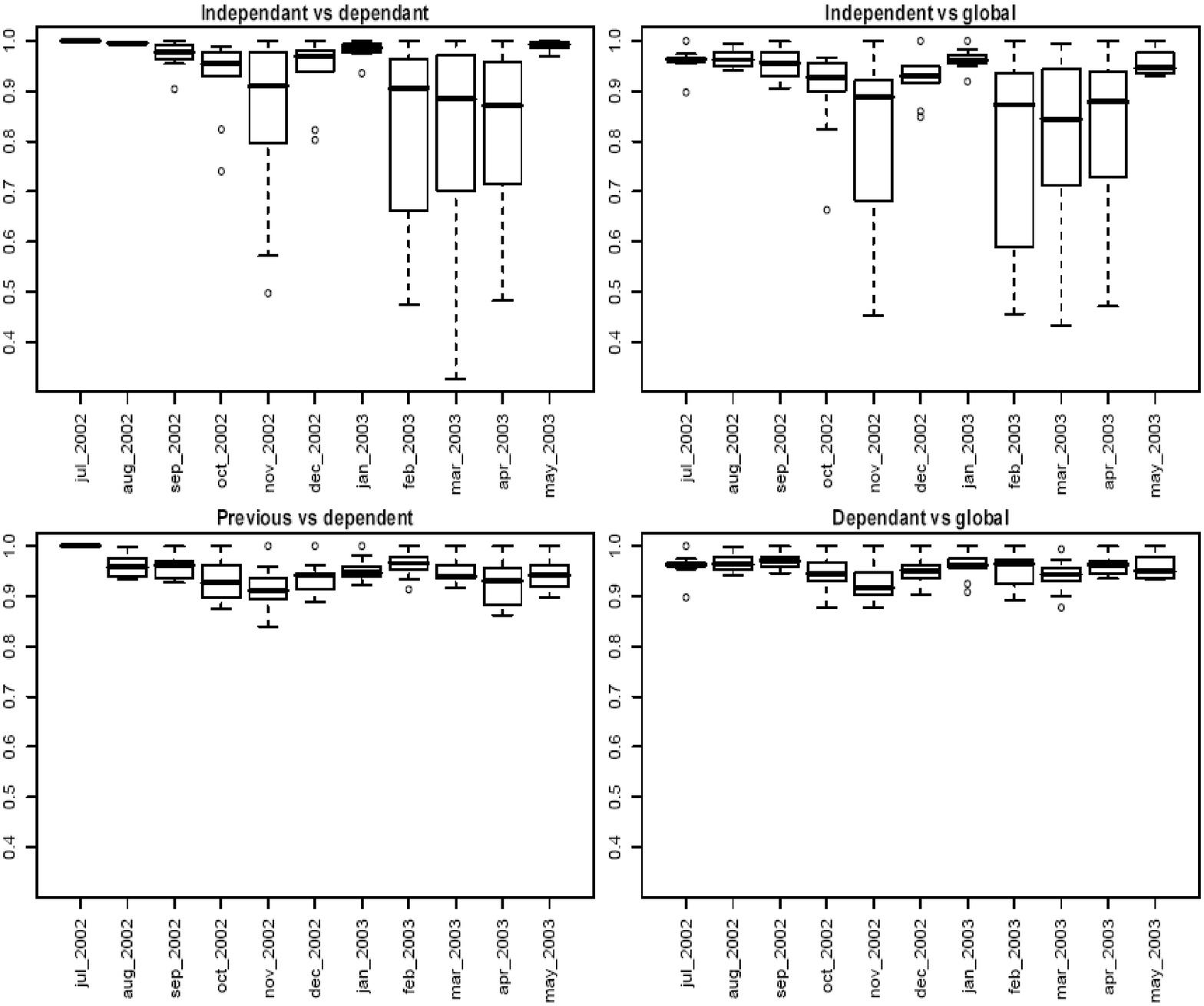}
\caption{F-measure values computed cluster by cluster} \label{fig_fmeasure}
\end{figure*}

Figure~\ref{fig_rand} depicts the CR values obtained from the confrontation of the resulting partitions reached by the four different clustering strategies presented in section \ref{sec:clusteringapproach}. For the F-measure, figure~\ref{fig_fmeasure} presents boxplots summarizing the 10 values obtained (one value per cluster) on each month analysed. The partitions compared are marked in red in figures \ref{fig_global_clustering}, \ref{fig_independent_local_clustering}, \ref{fig_previous_local_clustering} and \ref{fig_dependent_local_clustering}.

In figure~\ref{fig_rand}, the values of the CR index reveal that the results from the local independent clustering are very different from those of the local dependent and global clustering (low values of CR means that the partitions compared are divergent). 
In other words, we can see by the confrontation of the local independent clustering versus the local dependent and global clustering that there are almost always low values, i.e., the local independent clustering is able to find certain clusters that were not detected by the other two clustering strategies. These conclusions are also confirmed by the F-measure (cf. figure \ref{fig_fmeasure}).

We can also notice that the local previous clustering does not give very different results from those obtained by the local dependent clustering. The same conclusion is also valid for the confrontation between the local dependent clustering and global clustering.

Using a cluster-by-cluster confrontation via the F-measure, we refine the analysis (see figure \ref{fig_fmeasure}). What appears quite clearly is that the clusters are very stable over time if we apply either the local previous, local dependent or global clustering strategies. In fact, no value is lower than 0.877, which represents a very good score. On the other hand, in the case of local independent clustering, we detect clusters that are very different from those obtained by the global and local dependent clustering (some values are lower than 0.5). 

What is surprising is that partitions obtained by the local dependent clustering are very similar to those obtained from the global clustering. We could thus speculate whether an analysis carried out on time sub-periods would be able to obtain results supposed to be revealed by a global analysis on the entire data set. 

To summarize, we can say that the local dependent clustering method shows that the clusters obtained change very little or do not change at all, whereas the local independent clustering method is more sensitive to changes which occur from one time sub-period to another.

\section{Conclusion}

In the present work, we addressed the problem of processing dynamic data in the WUM domain. The issues discussed highlight the need to define or adapt methods to extract knowledge and to follow the evolution of this kind of data. Although many powerful knowledge discovery methods have been proposed for WUM, very little work has been devoted to handling problems related to data that can evolve over time. 

In this article, we have proposed a divide and conquer based approach in the sense that we split the original data set into temporal windows regarding the position of the data in the time scale. Our experiments have shown that the analysis of dynamic data by independent time sub-periods offers a certain number of advantages such as making the method sensitive to cluster changes over time. 

Analysing changes in clusters of usage data over time can provide important clues about the changing nature of how a website is used, as well as the changing loyalties of its users. Furthermore, as our approach splits the data and concentrates the analysis on fewer sub-sets, some constraints regarding hardware limitations could be overcome. 

Possible future work could involve the application of other clustering algorithms and the implementation of techniques that enable the automatic discovery of the number of clusters as well as identifying cluster fusions and splits over time.

\section*{Acknowledgements}
The authors are grateful to CAPES (Brazil) and the collaboration project between INRIA (France) and FACEPE (Brazil) for their support for this research.	
%
%
%
%
%

%
%



\printindex
\end{document}